%% file: acl_latex.tex
\pdfoutput=1

\documentclass[11pt]{article}

\usepackage[]{acl}

\usepackage{times}
\usepackage{latexsym}

\usepackage[T1]{fontenc}

\usepackage[utf8]{inputenc}

\usepackage{microtype}

\usepackage{inconsolata}

\usepackage{graphicx,xcolor}  
\usepackage{pxrubrica}        
\usepackage{fontawesome}
\usepackage{url}
\usepackage{comment}
\usepackage{multirow}
\usepackage{tikz}
\usepackage[xcolor,framemethod=tikz]{mdframed}
\usepackage{color}
\usepackage{bm}
\usepackage{booktabs}
\usepackage{amssymb}
\usepackage{subcaption}
\usepackage{siunitx}
\usepackage{colortbl}

\usepackage[whole]{bxcjkjatype}

\DeclareRobustCommand{\testbgfa}[1]{\tikz[baseline=(X.base)]{\node(X)[rectangle, fill=cyan!60!blue!18, rounded corners, text height=.8ex,text depth=-0.5ex]{\textit{#1}};}}
\DeclareRobustCommand{\testbgfb}[1]{\tikz[baseline=(X.base)]{\node(X)[rectangle, fill=red!60!blue!18, rounded corners, text height=.8ex,text depth=-0.5ex]{\textit{#1}};}}
\DeclareRobustCommand{\testbgfc}[1]{\tikz[baseline=(X.base)]{\node(X)[rectangle, fill=green!60!blue!18, rounded corners, text height=.8ex,text depth=-0.5ex]{\textit{#1}};}}

\newcommand{\query}{\{$query$\}}
\newcommand{\desc}{\{$description$\}}
\newcommand{\ads}{\{$ad text$\}}
\newcommand{\sysout}{\{$system$\}}
\newcommand{\reference}{\{$reference$\}}

\newcommand{\dataset}{\textsc{CAMERA}}

\title{Striking Gold in Advertising: Standardization and Exploration of Ad Text Generation}

\author{
  Masato Mita, 
  Soichiro Murakami, 
  Akihiko Kato, 
  Peinan Zhang \\
 \includegraphics[width=0.15\linewidth]{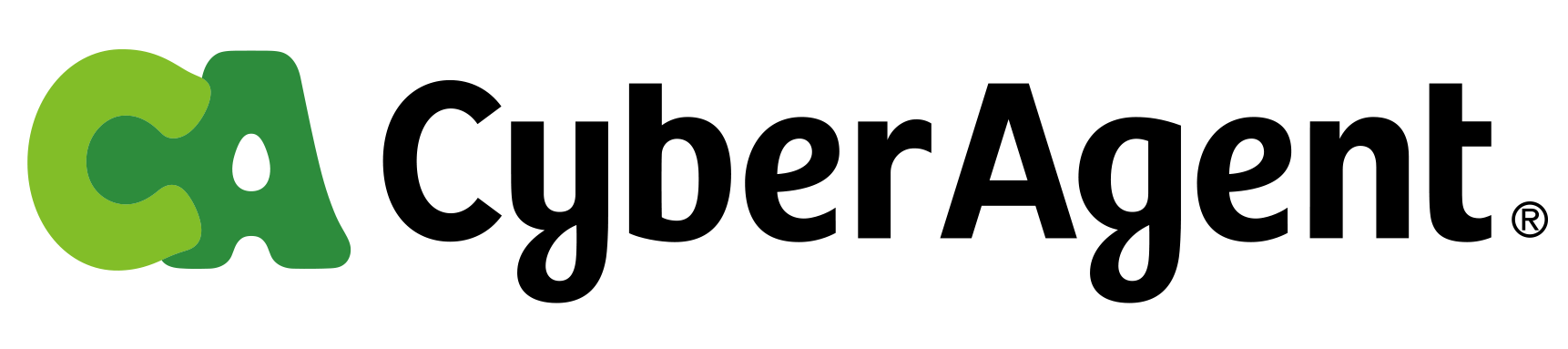} \\
 \{mita\_masato, murakami\_soichiro, kato\_akihiko, zhang\_peinan\}@cyberagent.co.jp\\
}

\begin{document}
\maketitle
\begin{abstract}
In response to the limitations of manual ad creation, significant research has been conducted in the field of automatic ad text generation (ATG). 
However, the lack of comprehensive benchmarks and well-defined problem sets has made comparing different methods challenging.
To tackle these challenges, we standardize the task of ATG and propose a first benchmark dataset, \dataset\faCameraRetro, carefully designed and enabling the utilization of multi-modal information and facilitating industry-wise evaluations.
Our extensive experiments with a variety of nine baselines, from classical methods to state-of-the-art models including large language models (LLMs), show the current state and the remaining challenges.
We also explore how existing metrics in ATG and an LLM-based evaluator align with human evaluations.
\end{abstract}

\section{Introduction}

\label{sec:intro}

The global online advertising market has witnessed significant growth and quadrupled over the last decade, particularly in the domain of search ads~\cite{Meeker:18}.
Search ads are designed to accompany search engine results and are tailored to be relevant to users' queries (search queries).
These ads are displayed alongside a landing page (LP), providing further details about the advertised product or service. 
Therefore, ad creators must create compelling ad texts that captivate users and encourage them to visit the LP.
However, the increasing volume of search queries, which is growing at a rate of approximately 8\% annually~\cite{Ogi:22}, poses challenges for manual ad creation.

\input{figures/fig_proposed_dataset}

The growing demand in the industry has fueled research on the automatic generation of ad texts. 
Researchers have explored various approaches, starting with \textit{template-based} methods that generate ad text by inserting relevant keywords into predefined templates~\cite{Bartz-et-al:08,Fujita:10,Thomaidou-etal_snippet:13}. 
Recently, neural language generation (NLG) techniques based on encoder-decoder models, which are widely employed in machine translation and automatic summarization, have been applied to ad text generation (ATG)~\cite{Hughes-et-al:19,Mishra-et-al:20,kamigaito-etal-2021-empirical}.

However, the automated evaluation of ATG models presents significant challenges.
Previous research has been constrained to conducting individual experiments using proprietary datasets that are not publicly available~\cite{murakami2023natural}.
This limitation arises from the absence of a shared dataset (i.e., a benchmark) that can be universally applied across the field.
Moreover, the absence of benchmarks has resulted in a lack of consensus regarding task settings such as the models' input/output formats.
While some studies use keywords as input~\cite{Bartz-et-al:08,Fukuda:19}, others employ existing advertisements~\cite{Mishra-et-al:20} or LPs~\cite{Hughes-et-al:19,kanungo-etal:22,golobokov-etal-2022-deepgen}. 
This variation in the task setting indicates that the field as a whole has yet to establish a standardized problem setting, which hinders the generalization and comparability of ATG techniques.

This study aims to advance ATG technology by standardizing the task setup, transforming it into a format accessible to potential players by providing a shared dataset, and exploring the current status and limitations.
Standardizing problem settings common to a variety of advertising applications as tasks allows for focused exploration of core issues in an academic context while maintaining the flexibility to be applied to a wide variety of applications (\S\ref{sec:definition}).
To engage a broader community of researchers beyond those who possess ad data, we construct the first publicly available benchmark, \textbf{\dataset\faCameraRetro} (\textbf{C}yber\textbf{A}gent \textbf{M}ultimodal \textbf{E}valuation for Ad Text Gene\textbf{RA}tion)(Figure~\ref{fig:camera_example}), which is meticulously developed a comprehensive dataset (\S\ref{sec:construction_data}).\footnote{\url{https://github.com/CyberAgentAILab/camera}; \url{https://huggingface.co/datasets/cyberagent/camera}} 
Our dataset comprises actual data sourced from Japanese search ads and incorporates annotations encompassing multi-modal information such as the LP images.
To explore the current state and future challenges, we conducted extensive experiments using nine diverse baselines, including multimodal models and large language models (LLMs), as well as the dominant approaches in existing studies (\S\ref{sec:experiments}).
Furthermore, we also conducted a meta-evaluation of how well the existing metrics and LLM-based evaluators reproduced human evaluations (\S\ref{sec:meta_evaluation}).

Our major contributions are:

\begin{itemize}
  \setlength{\itemsep}{0cm} 
    \item Establishing the standardized task and creating open data have paved the way for reproducible research and lowered barriers to entry.
    \item Benchmarking experiments with nine diverse models, including classical, standard, and state-of-the-art LLM-based models, demonstrated the current state and future challenges.
    \item The first meta-evaluation highlighted the reliability and limitations of automatic evaluations.
\end{itemize}

We observed the following:
\begin{itemize}
  \setlength{\itemsep}{0cm} 
    \item Fine-tuned encoder-decoder models play an important role in maximizing automatic evaluation scores and improving quality in intrinsic evaluations such as faithfulness and fluency.
    \item Few-shots with strong LLMs have great potential for quality improvement in extrinsic evaluations such as human preference.
    \item Using multimodal information like LP images improves ad quality, but methods for model integration require further exploration.
    \item Model performance and rankings vary by industry domain.
    \item Existing metrics work as intrinsic evaluations, but it is still difficult to use them as a substitute for extrinsic evaluations.
    \item Human preference serves as a rough estimation of performance values in online evaluation such as CTR.
\end{itemize}

\section{Background}
\label{sec:related_work}
Various types of online advertising exist, including search ads, display ads~\footnote{Display ads typically take the form of banner ads strategically placed within designated advertising spaces on websites or applications.}, and slogans~\footnote{Slogans are catchy phrases designed to captivate the attention of internet users and generate interest in products, services, or campaigns.}.
However, since most existing studies are related to search ads~\cite{murakami2023natural}, this study also focuses on search ads and provides an overview of ATG research and its current limitations.

\subsection{A quick retrospective}
\label{subsec:quick_retrospective}

Early ATG systems predominantly relied on template-based approaches~\cite{Bartz-et-al:08,Fujita:10,Thomaidou-etal_snippet:13}. 
These approaches involved filling appropriate words (i.e., keywords) into predefined templates, resulting in the generation of ad texts.
Although this method ensured grammatically correct ad texts, it has limitations in diversity and scalability because it could only accommodate variations determined by the number of templates, which are expensive to create.
To address these constraints, alternative approaches have been explored, including reusing existing promotional text~\cite{Fujita:10} and extracting keywords from LPs to populate template slots~\cite{Thomaidou-etal_snippet:13}.

Encoder-decoder models, which have demonstrated their utility in NLG tasks such as machine translation and summarization~\cite{Sutskever-etal:14}, have been applied to ATG research~\cite{Hughes-et-al:19,Youngmann-etal:20,kamigaito-etal-2021-empirical,golobokov-etal-2022-deepgen}. 
These models have been employed in various approaches, including \textit{translating} low click-through-rate (CTR) sentences into high CTR sentences~\cite{Mishra-et-al:20}, \textit{summarizing} crucial information extracted from the LPs~\cite{Hughes-et-al:19,kamigaito-etal-2021-empirical}, and combining these techniques by first summarizing the LPs and subsequently translating them into more effective ad texts based on CTR~\cite{Youngmann-etal:20}.\footnote{CTR is a widely-used indicator of advertising effectiveness in the online advertising domain.}
Recently, transfer learning approaches using pre-trained language models have become mainstream, allowing for more fluent and diverse ATG~\cite{wang2021reinforcing,Zhang-etal:21,golobokov-etal-2022-deepgen,kanungo-etal:22,wei-etal-2022-creater,li-etal-2022-culg,Murakami-et-al:22}.

\subsection{Current limitations}
\label{subsec:current_limitatioons}

ATG has experienced remarkable growth in recent years, garnering significant attention as a valuable application of natural language processing (NLP).
However, the automated evaluation of models presents substantial challenges.
Existing studies, validated only on \textit{non-public} datasets, hinder fair comparisons and discussions across studies, posing challenges in generalizing ATG technology.
Related to this, the problem settings for ATG, such as input/output, are not shared among the studies because there are variations depending on the advertising medium (e.g., search ads and display ads) and platform (e.g., Google and Bing).
These challenges are primarily due to the absence of a shared benchmark dataset that can benefit the entire research community.
The reason behind the reluctance to share ad datasets is that they usually contain performance values such as CTR, which are confidential data for companies.
Table~\ref{tab:summary_exisiting_work} summarizes the existing studies in the field and shows that this field is led by companies operating advertising-related businesses.
Moreover, it stands out as a valuable research subject contributing to the development of user-centered NLP techniques.
As a confluence of these trends, this study aims to establish ATG as an NLP task by standardizing the task and building a benchmark dataset.

\section{Standardization of ad text generation}
\label{sec:definition}
One of the goals of this study is to develop a task that is not specific to a particular platform or advertising medium but focuses on universal core problems common to these applications, to facilitate the generalization of ATG technology.
To meet these requirements, we standardize the ATG task as follows:
Let $\bm{x}$ be a source document that describes advertised products or services, $\bm{a}$ a user signal reflecting the user's latent needs or interests, and $\bm{y}$ an ad text.
ATG aims is to model $\bm{p(y|a,x)}$.
User signals, such as search keywords for search ads and user browsing and action history for display ads, can vary based on the application and domain. 
The specific data to be selected for each $\bm{x}$, $\bm{a}$, and $\bm{y}$ will be left to future dataset designers and providers.
This standardization of ATG allows a focused exploration of core issues in an academic context while maintaining flexibility for diverse applications in an industrial context.

\paragraph{The requirements of ad text}
The purpose of advertising is to influence consumers' (users) attitudes and behaviors towards a particular product or service. 
Therefore, the goal of ATG is to create text that encourages users' purchasing behaviors. 
Based on this, the following two requirements for ad text were defined:
(1) The information provided by the ad text is consistent with the content of the source document; and (2) the information is carefully curated and filtered based on the users' potential needs, considering the specific details of the merchandise.
Requirement 1 relates to \textit{hallucinations}, which is currently a highly prominent topic in the field of NLG~\cite{wiseman-etal-2017-challenges,parikh-etal-2020-totto,maynez-etal-2020-faithfulness}.
This requirement can be considered crucial for practical implementation since the inclusion of \textit{non-factual hallucination} in ad texts can cause business damage to advertisers.
Regarding requirement 2, it is necessary to successfully convey the features and attractiveness of a product within a limited space and immediately capture the user's interest. 
Therefore, ad text must selectively include information from inputs that can appeal to users.

\paragraph{Differences from existing tasks}
The ATG task is closely related to the conventional document summarization task in that it performs information compression while maintaining consistency with the input document's content.
Particularly, \textit{query-focused summarization (QFS)}~\cite{Dang:05}, a type of document summarization, is the closest in problem setting because it takes the user's query as the input; however, there are some differences.
The task of QFS aims to create a summary from one or multiple document(s) that answers a specific query (\textit{explicit needs}).
In contrast, ATG is required to extract not only surface information from user signals but also the \textit{latent needs} behind them and then return a summary.
For example, when a user's query is ``used cars,'' the goal of QFS is to provide information about used cars. 
On the other hand, for users seeking higher-priced items like cars, factors such as quality become important even if they are used. 
Therefore, the task of ATG aims to present ads that include expressions appealing to high quality and reassurance, such as \textit{``All cars come with a free warranty!''}.

Another notable difference is that while summarization aims to deliver accurate text that fulfills task-specific requirements, ATG surpasses mere accuracy and aims to influence user attitudes and behavior. 
Consequently, unconventional and/or ungrammatical text may be intentionally used in ad-specific expressions to achieve this objective (refer to details in \S\ref{subsec:flow}).
Therefore, QFS is a subset of ATG (QFS $\subset$ ATG).
One of the technical challenges unique to ATG is capturing users' latent needs based on such user signals $\bm{a}$ and generating appealing sentences that lead to advertising effectiveness, which depends significantly on the psychological characteristics of the recipient users.
Therefore, realizing more advanced ATG will also require a connection with advertising psychology~\cite{Scott:03} based on cognitive and social psychology.
The ATG is an excellent research topic for advancing user-centered NLP technologies.

\section{Construction of \dataset\faCameraRetro}
\label{sec:construction_data}

\subsection{Dataset design}
\label{subsec:design}
In this study, the following two design policies were first established: the benchmark should be able to (1) utilize multimodal information and (2) evaluate by industry domain.
In terms of~\textbf{Design Policy 1}, various advertising formats use textual and visual elements to communicate product features and appeal to users effectively. 
It is well-recognized that aligning content with visual information is crucial in capturing user attention and driving CTR.
\textbf{Design Policy 2} highlights the significance of incorporating specific~\textit{advertising appeals} to create impactful ad texts. 
In general, ad creators must consider various aspects of advertising appeals such as the \textit{price}, \textit{product features}, and \textit{quality}. 
For instance,  advertising appeals in terms of \textit{price} such as \textit{``get an extra 10\% off''} captivate users by emphasizing cost savings through discounts and competitive prices.
Previous studies revealed that the effectiveness of these advertising appeals varies depending on the target product and industry type~\cite{murakami-etal-2022-aspect}.

\subsection{Construction procedure}
\label{subsec:flow}

We utilized Japanese search ads from our company involved in the online advertising business.\footnote{We take great care to ensure that advertisers are not disadvantaged by the release of data.}
In these source data, the components of user queries, ad texts, and LPs (URLs) are allocated accordingly. 
Search ads comprise a~\textit{title} and \textit{description} as shown in Figure~\ref{fig:example_sea}.
Description in search ads has a larger display area compared to titles. 
It is typically written in natural sentences but may also include advertising appeals.
In contrast, titles in search ads often include unique wording specific to the advertisements. 
They may deliberately break or compress grammar to the extent acceptable to humans because their primary role is immediately capturing a user's attention.
For instance, the sentence \textit{``If you're looking to sell your brand-name merchandise, why not get a free valuation at XX right now?''} is transformed into an ad-specific expression: \textit{``Sell your brand-name goods / free valuation now''}.
Studies in advertising psychology have reported that these seemingly ungrammatical expressions, unique to advertisements, not only do not hinder human comprehension but also capture their attention~\cite{Wang-etal:13}.
We extracted only titles as ad texts $\bm{y}$ to create a benchmark focusing on ad-specific linguistic phenomena.

In our dataset, we extracted~\texttt{meta description} from the HTML-associated LPs, which served as a description document (\textit{LP description}) $\bm{x}$ for each product. 
Furthermore, in line with \textbf{Design Policy 1}, we processed a screenshot of the entire LP to obtain an LP image, allowing us to leverage multi-modal information. 
Through this process, we obtained images $\bm{I}$, layout information $\bm{C}$, and text~$\{x_i^{\texttt{ocr}}\} _{i=1}^{|\bm{R}|}$ for the rectangular region set $\bm{R}$ using the OCR function of the Cloud Vision API.\footnote{\url{https://cloud.google.com/vision/docs/ocr}}

\subsection{Annotation}

The source data is assigned a delivered gold reference ad text, but because of the variety of appeals in the ads, there is a wide range of valid references for the same product or service.
Therefore, three additional gold reference ad texts were created for the test set by three expert annotators who are native Japanese speakers with expertise in ad annotation.
The test set was obtained by randomly sampling about 1000 sentences (about 5\% of the total) of the source data set, considering the annotation cost and the need to ensure a minimum amount of data for evaluation purposes.\footnote{Excluded cases where LP URL was invalid after sampling.}
The detailed annotation guidelines are presented in Appendix~\ref{appendix:guidline}.
During the data collection process for evaluation annotations, data were randomly selected based on keywords manually mapped to industry labels, such as \textit{``designer jobs''} mapped to the human resource industry, following~\textbf{Design Policy 2}.
Here, we used the following four industry domain labels: human resources (HR), e-commerce (EC), finance (Fin), and education (Edu). 
The dataset was partitioned into training, development, and test sets to prevent data duplication between the training (development) and test sets, which was achieved through filtering processes.
\input{tables/tab_corpus_stats.tex}

Table~\ref{tab:corpus_stats} provides the statistics of our dataset.
It is worth noting that more information can be taken into account, including not only the text information (\textit{LP desc.}) of the LP, but also the text written on the image by applying OCR processing to the LP image (\textit{LP OCR}).
Figure~\ref{fig:camera_example} presents examples from the test set of this dataset.
Although the annotator was not given explicit instructions regarding the advertising appeal, we confirmed that the annotator created an ad text (\#2-4) that featured a variety of advertising appeals different from the original ad text (\#1) that considered latent needs based on keywords.
This suggests that our test set captures a certain level of diversity in expressing advertisements.
To emphasize the multimodal nature of this dataset, we provide examples of ad texts that are difficult to generate without understanding the LP's image information in Appendix~\ref{appendix:example_ads_multimodal}.

\input{tables/tab_entity-type}

\input{figures/fig_camera-train_hallucination}
\subsection{Understanding of human ad creation}
To gain more insight into the dynamics of human ad creation, we investigated the extent to which ad creators are making their ads extractive (or abstractive).
This exploration would be also useful as a guideline for future model development.

Figure~\ref{fig:camera-train_hallucination} illustrates the percentage of \textit{novel} entities in the target ad texts not found in their respective source documents. 
Here, we focused on five distinct entity types as outlined in Table~\ref{tab:entity_type} to conduct a more comprehensive analysis.\footnote{The procedure for calculating the ratio of novel entities is described in Appendix~\ref{appendix:novel_entity}.}
By incorporating additional input information such as the LP description and OCR-processed text of the LP full view, the percentage of novel entities in the target ad text was effectively reduced.
Furthermore, the analysis based on entity type reveals a wide range of variations in \textit{Time Expressions} and \textit{Numerical Expressions}. 
In the example of~\textit{Numerical Expressions} as shown in Table~\ref{tab:entity_type}, the source document $x$ mentioned the price range as \textit{6,800 yen - 8,000 yen}, while the target ad text $y$ only included the lower limit of the range as \textit{6,800 yen}.
This rewording may be intended to make the price more appealing to users by presenting the lowest price, or to make it more straightforward to fit into a limited display area.

\section{Benchmarking of ATG models}
\label{sec:experiments}

To clarify the current state and remaining challenges, we conduct benchmark experiments using the dataset constructed in \S\ref{sec:construction_data} and various ATG models. 
Specifically, we investigate the following research questions: 

\begin{itemize}
  \setlength{\itemsep}{0cm} 
 \item[\textbf{RQ1}] \textit{How do differences in the use of pre-trained language models (i.e., finetuning vs. few-shot) affect overall performance?}

    \item[\textbf{RQ2}] \textit{Is multimodal information useful for ad text generation?}

    \item[\textbf{RQ3}] \textit{Do trends in model performance vary by industry domain?}

    \item[\textbf{RQ4}] \textit{What are the qualitative differences between generated ad text compared to human-produced ad text?}
\end{itemize}

\subsection{Models}
\label{subsec:models}
As outlined in \S\ref{subsec:current_limitatioons}, existing studies use non-public data with performance values, such as CTRs, and therefore cannot be replicated on the \dataset~data set, which does not include performance values.
Therefore, this experiment will focus on a simplified replication of previous studies and follow-up on the dominant approach.

\begin{itemize}
  \setlength{\itemsep}{0cm} 
    \item \textbf{BM25}
    is a model of an extractive approach using the BM25 algorithm~\cite{robertson2009probabilistic}. The BM25 algorithm is used to generate ad texts by extracting one query-related sentence from the input document. 

    \item \textbf{BART}
    is a fine-tuned model using BART~\cite{lewis-etal-2020-bart}. We used the following pre-trained model: \texttt{japanese\_bart\_base\_2.0}~\footnote{\url{https://github.com/utanaka2000/fairseq/tree/japanese_bart_pretrained_model}}
    \item \textbf{T5}~is a fine-tuned model using T5~\cite{raffel-etal-2022-t5}. We used the following pre-trained model: \texttt{sonoisa/t5-base-japanese}~\footnote{\url{https://huggingface.co/sonoisa/t5-base-japanese}}.
    
    \item \textbf{GPT-3.5} is a few-shot model using GPT-3.5 (\texttt{gpt-3.5-turbo-0613})~\cite{Ouyang0JAWMZASR22}. We built the model using the API provided by OpenAI~\footnote{\url{https://github.com/openai/openai-python}}. 

    \item \textbf{GPT-4} is a  few-shot model using GPT-4 (\texttt{gpt-4-0613})~\cite{openai2023gpt4}. As with GPT-3.5, we constructed the model using the API provided by OpenAI.
    
    \item \textbf{Llama2} is a few-shot model using Llama2~\cite{touvron2023llama}. We used the following pre-trained model: \texttt{ELYZA-japanese-Llama-2-7b-instruct}~\footnote{\url{https://huggingface.co/elyza/ELYZA-japanese-Llama-2-7b}}.
\end{itemize}
For BART and T5, we fine-tuned each pre-trained model on the train split of \dataset.
For GPT-3.5, GPT-4, and Llama2, the baseline models were constructed by 3-shot in-context learning, respectively. 
To investigate the effectiveness of incorporating multi-modal features such as images and layout in the LPs and their impact on the overall performance, we built various settings for the T5-based model that considered LP image information, following \citet{Murakami-et-al:22}. Specifically, we incorporated the following three types of multi-modal information into the model architecture: LP OCR text (\texttt{lp\_ocr;o}), LP layout information (\texttt{lp\_layout;l}), and LP BBox image features (\texttt{lp\_visual;v}).
See Appendix~\ref{appendix:detailed_setting} for details on the experimental setup for each baseline model, including the prompt template.

\subsection{Evaluation}
\label{subsec:evaluatioin}

\paragraph{Automatic evaluation}
To evaluate the generated texts quality, we employed two widely used metrics in ATG: BLEU-4 (B-4)\footnote{\url{https://github.com/mjpost/sacrebleu}}~\cite{papineni-etal-2002-bleu} and ROUGE-1 (R-1)~\cite{lin-2004-rouge}. 
These metrics assess the similarity between the generated text and reference based on $n$-gram overlap. 
Since paraphrases are commonly used in ad texts, BERTScore (BS)~\cite{bertscore}, an embedding-based metric, was also used to handle their semantic similarity.
Additionally, as task-specific guardrails, we introduce keyword insertion rates (\textsc{Kwd})~\cite{Mishra-et-al:20} and sentence length regulation compliance rates (\textsc{Reg}).
\textsc{Kwd} represents the percentage of cases where the specified keyword is included in the generated text for evaluating the relevance of the LP and the ad text.
\textsc{Reg} indicates the percentage of compliance with the character count regulation (15 characters or less).

\paragraph{Manual evaluation}
To answer RQ4, we conducted a manual evaluation.
Three human raters who are native Japanese speakers with expertise in ad annotation evaluate each of the 10 ad texts of the 9 models (\S\ref{subsec:models}) and one original reference for each of the three evaluation aspects of \textit{faithfulness}, \textit{fluency}, and \textit{attractiveness}.
The faithfulness and fluency evaluations were conducted using an \textit{absolute} evaluation of whether the input document implies or does not imply the ad text, and whether the content of the ad text is understandable and natural, respectively.
Given the challenge of providing an absolute evaluation of each ad text's attractiveness, we conducted a pairwise evaluation comparing the human reference and each model output, considering cases where the attractiveness was equal (\textit{Tie}).
For faithfulness and fluency, we sampled 200 cases from the test data and conducted manual evaluations for a total of 2000 ad texts. 
For attractiveness, we sampled 100 cases, created pairs of the human reference and each model output, and performed manual evaluations for a total of 900 ad texts.
Details of the instructions in the manual evaluation are provided in Appendix~\ref{appendix:detailed_setting_human-eval}.

Table~\ref{tab:agreement} shows the inter-annotator agreement (IAA)\footnote{It is based on majority vote and counted as a Tie if they are all split for attractiveness}. 
As expected, the IAA for attractiveness is the lowest, but when loosened to more than a majority, it is outstandingly high (0.84).
This suggests that, while achieving unanimous favorability is challenging, there is a considerable level of consensus on attractiveness.

\input{tables/tab_agreement}

\input{tables/tab_main_exp}
\input{figures/fig_industry_wise_eval}
\input{figures/fig_human-eval_faithful-fluency}
\input{figures/fig_human-preference}
\input{figures/fig_multimodal_example}

\subsection{Result}
The answers corresponding to the RQs listed in \S\ref{sec:experiments} are provided below:

\paragraph{A1: Finetuning and few-shot are good performers in intrinsic and extrinsic evaluations, respectively}
In automatic evaluation, we observe that few-shot learning falls behind finetuning (Table~\ref{tab:main_results}).
A similar trend can also be observed in the manual evaluation, except for attractiveness (Figure~\ref{fig:main_faithful_fluency} and Figure~\ref{fig:attractivenss}).
These series of results highlight the high potential of LLM few-shot for improving quality in \textit{extrinsic} evaluation such as attractiveness and human preference, while finetuning can play an important role in maximizing quality in \textit{intrinsic} evaluation such as automatic scores, faithfulness, and fluency.

\paragraph{A2: Multimodal information contributes to the quality of generated ad text}
We observe that incorporating additional features such as OCR-processed text (+ \texttt{\{o\}}), the LP layout information (+ \texttt{\{o,l\}}), and LP image features (+ \texttt{\{o,l,v\}}) improved the quality of generated sentences in terms of faithfulness (\ref{fig:faithful}) and fluency (\ref{fig:fluency}).
On the other hand, the incorporation of layout information and visual features into the models does not necessarily improve performance, so methods for model integration require further exploration.
Nevertheless, we also confirmed cases where the use of multimodal information in LPs improves the quality of the generated ad text as shown in Figure~\ref{fig:multimodal_example}.
The performance drop may be due to image information acting as noise when using the LP Full View directly in this experiment.
Therefore, the development of a multimodal system that adaptively accesses only important information from LPs will be a straightforward future work.

\paragraph{A3: Model performance and model rankings vary by industry domain}
Figure~\ref{fig:industry_wise_eval} shows the industry-wise evaluation results in each metric\footnote{We provide the details of the results in Appendix~\ref{appendix:detailes_industry}}.
We observe the model performance and rankings vary by industry.
This suggests that the performance of ATG models is sensitive to the industry domain and highlights the need for industry-wise evaluation to develop robust models.

\paragraph{A4: Some baselines have already reached human-level performers}
In \textbf{faithfulness}, the outputs of the baseline models, except GPT-3.5 and GPT-4, are more faithful to the input than the human reference (Figure~\ref{fig:faithful}).
Note, however, that low faithfulness in human reference does not necessarily mean low quality, since it is known that ad creators use expressions based on their external knowledge to the extent that they can ensure factual consistency with the input to enhance fluency and appeal.
Non-factual, fake ads can be fatal to advertisers in terms of legal compliance and corporate branding, but it is difficult for a model to perfectly capture real-time product-specific information, such as discount prices and campaign periods.
Therefore, one important direction is the development of models with guaranteed faithfulness as a step toward achieving an ATG system with guaranteed factual consistency.

In \textbf{fluency}, we can confirm that the human reference has high fluency as a trade-off for low faithfulness, while GPT-4, T5, and Llama2 are almost at the same level as the human reference (Figure~\ref{fig:fluency}).
It should also be noted that integrating multimodal information from LP images into the model contributes to generating more fluent ad text.

In \textbf{attractiveness}, GPT-4 is already able to generate more attractive ad text for humans than reference (Figure~\ref{fig:attractivenss}).
If equivalent (Tie) cases are included, T5 and T5+ \texttt{\{o\}} also reach the same level as humans.
GPT-4 also achieves a sentence-length regulation compliance rate (\textsc{Reg} in Table~\ref{tab:main_results}), making it a model with high real-world applicability.

\input{tables/tab_meta_eval}
\input{figures/fig_kiwami_pref}

\section{Analysis}

\subsection{How well can automated evaluations replicate human evaluations?}
\label{sec:meta_evaluation}
To clarify the limitations and possibilities of automatic evaluation, we performed a meta-evaluation by adding a GPT-4 based evaluator to the set of the metrics used in the experiment in \S\ref{sec:experiments}. 
The GPT-4 based evaluator was constructed by giving the same instructions as those given to the human raters in the manual evaluation~\S\ref{subsec:evaluatioin}.\footnote{The prompts used are presented in Appendix~\ref{appendix:gpt4_evaluator_prompts}.}

Table~\ref{tab:meta_eval} shows that the system-level meta-evaluation results with Pearson (r) and Spearman ($\rho$).
BS and R-1 correlate best with humans for faithfulness and fluency, respectively. 
On the other hand, it was difficult to replicate the human ranking for attractiveness.
This suggests that existing metrics work as intrinsic evaluations, but it is still difficult to use them as a substitute for extrinsic evaluations.
The GPT-4 based evaluator had the lowest correlation in any evaluation aspect.
This result is inconsistent with the existing studies~\cite{chiang-lee-2023-large,zheng2023judging}'s report that LLM evaluations produce results similar to those of expert human evaluations.
One reason for this may be due to domain mismatch, as most of the datasets in the GPT-4 pre-training are general or non-advertising domains~\cite{openai2023gpt4}.

\subsection{How well does human preference align with advertising performance?}
\label{subsec:kiwami}
To clarify the extent to which the human preference in Figure~\ref{fig:attractivenss} is aligned with advertising performance such as CTR, we investigate the rate of agreement between human preference and CTR.
Measuring CTR requires deploying the system output obtained in \S\ref{sec:experiments} as online advertisements, which is impractical. 
Therefore, we follow methodologies established in previous studies\cite{Rennie_2017_CVPR,Hughes-et-al:19} and approximate it using a CTR prediction model (i.e., predicted CTR; \textit{pCTR}).\footnote{We utilized 極予測TD (Kiwami Yosoku TD), our company's off-the-shelf model for pCTR calculation, which aligns with CTR. cf. \url{https://cyberagent.ai/products/}}

Table~\ref{fig:kiwami_pref} shows the agreement rate between human preference and pCTR when divided into bins according to the size of the ratio of pCTR (henceforth, \textit{pCTR ratio}) between reference and system, which is calculated as pCTR ratio = pCTR  (system) / pCTR (reference).
The results suggest that as the pCTR ratio decreases (indicating greater expected effectiveness of the reference over the system), humans find the reference more appealing. Conversely, when the pCTR ratio exceeds 1.0 (indicating the system outperforms the reference in expected advertising effectiveness), human preference and pCTR are less likely to align.
This suggests that in the band of performance where there's room for improvement, indicated by the generated ad text quality falling below the human reference, leveraging human preferences as an estimate of ad performance values like CTR is effective. Conversely, as the quality of the generated ad text approaches saturation and surpasses the human reference, it's advisable to incorporate online evaluation such as CTR measurement alongside offline evaluation to verify advertising effectiveness.

\section{Discussion for reproducible research}
We want to situate our findings in the context of the broader NLP community, in line with our goal of discussion on increased transparency in the field.
Examples of data that are challenging to open include proprietary datasets primarily owned by companies, housing sensitive information for maintaining a competitive advantage (e.g., datasets managed by OpenAI).
Ad data, the primary focus of this study, also exemplified this scenario.
One of the reasons why ad data has not been shared with the community in the past is that CTRs and other performance data are confidential.
Also, measuring CTR is difficult except for a few companies in the advertising business.
Therefore, as an incentive mechanism to promote the creation of open research within the research community, there is a direction for the community to accept secondary information (e.g., pCTR or human preference in \S\ref{subsec:kiwami}) that is guaranteed to be consistent to some extent with sensitive primary data (e.g., CTR).

\section{Conclusion}
We standardized ATG as a cross-application task and developed the first benchmark dataset. 
Through evaluation experiments using our dataset, we demonstrated the current status and remaining challenges.
ATG is a promising application of NLP and a critical and complex research area for advancing user-centric language technology. 
We hope that the research infrastructure we established will drive the progress and development of ATG technology.

\section*{Acknowledgements}
We thank the anonymous reviewers for their helpful comments and suggestions. 
We are also thankful to Ukyo Honda, Shota Sasaki, Sho Hoshino and the other members of CyberAgent for their insightful comments and suggestions.

\section*{Limitations}
One of the limitations of this study is that the dataset is only available in Japanese.
In particular, the community should also enjoy benchmark datasets in English that are more accessible to researchers and developers around the world.
We hope that advertising-related companies who share our vision of building on common datasets to build on the technologies in the field of ATG will follow this research and provide public datasets to the community for reproducible NLP research.

\bibliography{anthology,custom}

\appendix

\section{Example of search ads}
We provide an example of search ads in Figure~\ref{fig:example_sea}.

\section{Summary of existing studies}
\input{figures/fig_example_sea}

\input{tables/tab_exisiting_work_summary}

A summary of existing studies of ad text generation is shown in Table~\ref{tab:summary_exisiting_work}. 
From this, we can see that (1) the field is primarily led by companies related to the online advertising business, (2) there is no consensus on inputs and outputs, and (3) research has begun to flourish in the ACL community in recent years.

\section{Annotation guideline}
\label{appendix:guidline}
The main instructions given to the annotators were as follows:
\begin{enumerate}
    \item Consider the search keyword as the user's intent.
    \item Create an advertisement that is consistent with the product/service description in the LP.
    \item Ensure that the length of the advertisement is within 15 full-width characters~\footnote{This follows the guidelines for headline text in Google Responsive Search Ads (\url{https://support.google.com/google-ads/answer/12437745}).}.
    \item  These instructions were provided to guide the annotators in creating the additional reference advertisements.
\end{enumerate}

As explained in \S\ref{sec:definition}, since it is important for ad creation to consider latent needs behind user signals, we instructed the annotators to explicitly consider search keywords as user intentions.

\section{Examples of ad texts that are difficult to generate without considering LP image information}
\label{appendix:example_ads_multimodal}

\input{tables/tab_example_ads_multimodal}

\input{figures/fig_example_LP.tex}

We provide an actual example observed in our dataset in Table~\ref{tab:example_ads_multimodal}, which is difficult to generate without considering LP image information.
Based on the hypothesis that advertisements generally use visual information in addition to text to more effectively promote their products to users, we decided to include LP image information in the data set (i.e., Design Policy 1 in \S\ref{subsec:design}). 
In the dataset we constructed, we observed that the LP description alone was not sufficient and that the ad texts text required a deep understanding of the textual information embedded and the table data in the LP image.

\section{Calculation of ratio of novel entities}
\label{appendix:novel_entity}
In this section, we describe the procedure for calculating the ratio of novel entities, that is, entities that appear only on the output side. The target entity types are the following five types: (1) named entities, (2) terms, (3) katakana, (4) time expressions, and (5) numerical expressions. Let x denote the input and y the output ad sentence. Then, a procedure for calculating the ratio of novel entities is described as follows.

\begin{enumerate}
\item We perform NFKC normalization and lowercasing for each sentence in x and y.
\item In each instance (x,y), we perform the following (a)-(c) for each entity type $t_i$.

(a) We extract entity mentions of type $t_i$ from x and y (we call them $S_x^{(i)}$ and $S_y^{(i)}$, respectively) \footnote{We use GiNZA in spacy (ja\_ginza) for named entities, pytermextract (~\url{ http://gensen.dl.itc.u-tokyo.ac.jp/pytermextract}) for term extraction, regular expression for katakana, ja\_timex (~\url{https://github.com/yagays/ja-timex}) for time expressions, and pynormalizenumexp(~\url{https://pypi.org/project/pynormalizenumexp}) for numerical expressions.}. Regarding time expressions, we perform not only entity mention extraction, but also entity linking (e.g., "decade" and "10 years" are linked to the same time expression entity) thanks to the ja-timex library.

(b) We get novel entity mentions $S_{novel}^{(i)}$ by the following procedure. The following (i) or (ii) is used as the criterion to judge whether a given entity mention is novel or not. (i) In the case of the perfect match criterion (for katakana, time expressions, and numerical expressions): $S_{novel}^{(i)} = S_y^{(i)} / S_x^{(i)}$ (ii) For partial matching criteria (for named entities and terms): if each mentions in $S_y^{(i)}$ is not a full or partial match with any mentions in $S_x^{(i)}$, it is judged as a novel entity mention and added to $S_y^{(i)}$. As for named entities and terms, we adopt the partial matching criterion (ii), because there are many cases in which most of the entity mentions are identical, such as "Sendai" and "Sendai-city" in our initial exploration for sampled 100 instances from our dataset.

(c) If $S_y^{(i)}$ is not an empty set, then the ratio of novel entities for type $t_i$ for a given instance is calculated by $|S_{novel}^{(i)}| / |S_y^{(i)}|$.
\item Finally, for each entity type $t_i$, we compute the macro-averages of the above ratios for the set of instances in which entity mentions of type $t_i$ occur at least once in y.
\end{enumerate}

\section{Details on experimental setup for each baseline models}
\label{appendix:detailed_setting}

\subsection{BM25}
We used the BM25 to rank sentences of the source document given a query and took the most relevant sentence as the generated ad text. 
For implementation, we used the \texttt{rank\_bm25} toolkit~\footnote{\url{https://github.com/dorianbrown/rank\_bm25}}.

\subsection{T5 and BART}
We fine-tuned each pre-trained model on the training dataset to create our baseline models.
Specifically, we used a pre-trained model \texttt{japanese\_bart\_base\_2.0} from Kyoto University's Japanese version of BART~\footnote{\url{https://github.com/utanaka2000/fairseq/tree/japanese_bart_pretrained_model}} as the basis for our BART-based baseline model. 
For the T5-based baseline model, we used a pre-trained model \texttt{sonoisa/t5-base-japanese}~\footnote{\url{https://huggingface.co/sonoisa/t5-base-japanese}}.
The specific hyperparameters and other experimental details are reflected in Table~\ref{tab:hyper_parameter}.

\subsection{Multimodal models}
\label{appendix:lp_to_text}
\input{figures/fig_lp-to-text}

Figure~\ref{fig:lp-to-text} presents an overview of incorporating the LP information into the T5-based model.
\footnote{Note that the model constructed for this experiment, shown in Figure~\ref{fig:lp-to-text}, is not the proposed model, but a baseline model created according to \citet{Murakami-et-al:22}}.
As an input, we used three sets of token sequences, the LP descriptions $x^{des}$, user queries $x^{qry}$, and each OCR token sequence $x_i^{ocr}$ of the rectangular region set $R=\{r_i\}_{i=1}^{|R|}$ obtained by OCR from the LPs, where each token sequence $x^{*}$ is $x^{*}= (x_i^{*})_{t=i}^{|R|}$. 
Furthermore, the layout $C={c_{i}}_{i=1}^{|R|}$ and image information $I={I_{i}}_{i=1}^{|R|}$ for the rectangular region set $R$ was used.
Here, $c_i$ denotes $(x_i^{\mathsf{min}},x_i^{\mathsf{max}},y_i^{\mathsf{min}},y_i^{\mathsf{max}})\in\mathbb{R}^4$ as shown in Figure~\ref{fig:lp-to-text}.

Next, we explicitly describe each embedding (Figure~\ref{fig:lp-to-text}) as follows:
\paragraph{Token embedding}
Each token sequence $x^{*}$ was transformed into an embedding sequence $t^{*}$ before being fed into the encoder. 
Here, $D$ denotes the embedding dimension.

\paragraph{Segment embedding}
The encoder distinguishes the region of each token sequence $x^{*}$.
For example, for a token sequence $x^{des}$, we introduced $s^{des} \in \mathbb{R}^{D}$.

\paragraph{Visual embedding}
We introduced an image $I_{i}$ for each rectangular region $r_{i}$ to incorporate visual information from the LP, such as text color and font.
More specifically, the obtained image $I_{i}$ was resized to 128 $\times$ 32 (width × height). 
The CNN-based feature extraction was employed to create visual features $v_{i} \in \mathbb{R}^{D}$.

\paragraph{Layout embedding}
In the LP, the position and size of the letters played crucial roles. 
We input the layout $c_{i}$ of a rectangular region $r_{i}$ into the MLP to obtain $l_{i} \in \mathbb{R}^{D}$.

Using the above embeddings, we generated the encoder inputs, as shown in Figure~\ref{fig:lp-to-text}.
This study investigated the contribution of each type of multi-modal information to the overall performance.
We incorporated the following three types of multi-modal information into the model architecture in Figure~\ref{fig:lp-to-text}: LP OCR text (\texttt{lp\_ocr;o}), LP layout information (\texttt{lp\_layout;l}), and LP BBox image features (\texttt{lp\_visual;v}). 

\paragraph{Hyperparameters}
We present the hyperparameters used during the training of both models in Table~\ref{tab:hyper_parameter}. For the maximum sequence length in T5, it was set to 712 only for the model using LP bounding box image features (+ \texttt{\{o,l,v\}}), while all other models were set to 512. Furthermore, early stopping was applied if the loss on the development set deteriorated for 3 consecutive epochs in the case of T5, and 5 consecutive epochs in the case of BART.
\input{tables/tab_hyperparameters.tex}

\subsection{GPT-3.5, GPT-4, and Llama2}
For GPT-3.5, GPT-4, and Llama2, the baseline models were constructed by 3-shot in-context learning, respectively. 
The prompts used to build these models are provided in Table~\ref{tab:prompt_llm}.

\begin{table*}[t]
\small
\centering
\begin{tabular}{p{0.8\linewidth}}
\toprule[.1em]
Based on the given search query and text, please create an advertisement that appeals to users in 15 words or less.\\
\\
Search Query: bridal fair Yokohama \\
Document: Official website of "The House Yokohama Marine Tower Wedding", a wedding venue at Yokohama Marine Tower adjacent to Yamashita Park. One couple can rent out the Yokohama Marine Tower, which overlooks Minato Mirai, and have a wedding ceremony that is unique to them. \\
Output: Yokohama wedding THE HOUSE open \\
\\
Search Query: window cleaning \\
Documents: Compare window and sash cleaning prices, quotes, and reviews at Kurashi no Market. Easily book reputable window and sash cleaning professionals online! [Guaranteed!] \\
Output: [Official] Kurashino Market \\
\\
Search Query: jobs osaka 50s \\
Documents: Find the right job for you at Recruit's job search and job information site! Rikunabi NEXT is a job search and recruitment information site that supports your job search with useful content such as job scout function and know-how on job change. \\
Output: Many senior jobs are available \\
\\
Search query: \query \\
Documentation: \desc \\
Output:  \\
\bottomrule[.1em]
\end{tabular}
\caption{\label{tab:prompt_llm}
Prompts used for the ATG model based on LLMs (GPT-3.5, GPT-4, and Llama2), were translated into English for visibility.
}
\end{table*}

\section{Details on experimental setup for manual evaluation}
\label{appendix:detailed_setting_human-eval}
\input{figures/fig_faithfulness}
\input{figures/fig_fluency}
\input{figures/fig_attractivenss}

Three native Japanese speakers and advertising annotation experts were recruited from our in-house annotation center.
As an overview of the annotation, we instructed that a human evaluation of the quality of the ad text be conducted for each of the following three evaluation perspectives.

\begin{itemize}
     \item \textbf{faithfulness:} \textit{Does the LP description imply the ad text?}
    \item \textbf{fluency:} \textit{Is the content understandable and natural as an ad text?}
    
    \item \textbf{attractiveness:} \textit{Is it an attractive ad text?}
\end{itemize}

When conducting the annotation, the following tasks were created for each evaluation perspective: faithfulness (Figure~\ref{fig:faithul_anno}), fluency (Figure~\ref{fig:fluency_anno}), and attractiveness (Figure~\ref{fig:attractivenss_anno}), using \texttt{Label Studio}\footnote{\url{https://labelstud.io/}}, an open-source annotation tool.\footnote{All task examples are translated into English for visibility.}
Since faithfulness and fluency are absolute evaluations, 10 ad texts (9 systems + 1 reference) were evaluated together as shown in Figure~\ref{fig:faithul_anno} and Figure~\ref{fig:fluency_anno} to reduce annotation costs.
To ensure the quality of the annotations, a training phase was established before the test phase, and the annotators were trained with a total of 60 ad texts, 20 for each task.
The entire annotation process took roughly 6 hours.

\section{Prompts for GPT-4 evaluator}
\label{appendix:gpt4_evaluator_prompts}
The GPT-4-based evaluator was constructed by giving the same instructions as those given to the human raters in the manual evaluation~\S\ref{subsec:evaluatioin}. 
We present the prompts we used for faithfulness, fluency, and attractiveness in Tables~\ref{tab:prompt_gpt4_faithfulness}, Table~\ref{tab:prompt_gpt4_fluency}, and Table~\ref{tab:prompt_gpt4_attractiveness}, respectively.

\begin{table*}[t]
\small
\centering
\begin{tabular}{p{0.8\linewidth}}
\toprule[.1em]
Please answer "1" if the question text implies the ad text and "0" if it does not.\\
\\
Question text: [A calm daily life begins with a regular diet] Self-care for common female problems/regular delivery costs about 81 yen a day. \\
Ad text: Peaceful everyday life \\
Answer: 1 \\
\\
Question text: [A calm daily life begins with a regular diet] Self-care for common female problems/regular delivery costs about 81 yen a day. \\
Ad text: [Official] Daily diet \\
Answer: 0 \\
\\
Question: How to recover/restore data from an external HDD? \\
Ad text: 0 yen for the initial cost \\
Answer: 0 \\
\\
Question: \desc \\
Ad text: \ads \\
Answer:  \\
\bottomrule[.1em]
\end{tabular}
\caption{\label{tab:prompt_gpt4_faithfulness}
Prompt used for GPT-4 evaluator for faithfulness, translated into English for visibility.
}
\end{table*}

\begin{table*}[t]
\small
\centering
\begin{tabular}{p{0.8\linewidth}}
\toprule[.1em]
Please answer "1" for the following ad text if the content is understandable and natural, and "0" otherwise.\\
\\
Ad text: You get muji miles every year. \\
Answer: 1 \\
\\
Text: [Official] marriveil \\
Answer: 1 \\
\\
Ad text: ujipassport app \\
Answer: 0 \\
\\
Ad text: \ads \\
Answer:  \\
\bottomrule[.1em]
\end{tabular}
\caption{\label{tab:prompt_gpt4_fluency}
Prompt used for GPT-4 evaluator for fluency, translated into English for visibility.
}
\end{table*}

\begin{table*}[t]
\small
\centering
\begin{tabular}{p{0.8\linewidth}}
\toprule[.1em]
Assuming a Google search for the following keywords, please compare ad text A and ad text B and answer "A" or "B" for the one you are more interested in.
If the attractiveness is the same, please answer "C".\\
\\
Keyword: employment information \\
Ad text A: [Official] TOYOTA / Recruitment of periodic employees \\
Ad text B: [Official] TOYOTA / Periodic Employee Recruitment \\
Answer: A \\
\\
Keyword: recommended medical insurance \\
Ad text A: Nippon Life Group Medical Insurance \\
Ad text B: Online Medical Insurance \\
Answer: B \\
\\
Keyword: cancer hospital visit insurance \\
Ad text A: Sony Assurance's medical insurance \\
Ad text B: Aflac medical insurance \\
Answer: C \\
\\
Keyword: \query \\
Ad text A: \reference \\
Ad text B: \sysout \\
Answer: \\
\bottomrule[.1em]
\end{tabular}
\caption{\label{tab:prompt_gpt4_attractiveness}
Prompt used for GPT-4 evaluator for attractiveness, translated into English for visibility. The examples of prompts were selected by sampling from cases in which the evaluators' opinions were in total agreement during the manual evaluation.
}
\end{table*}

\section{Details of industry-wise evaluation}
\label{appendix:detailes_industry}

\input{tables/tab_industry-wise}

Details of the results of the evaluation of the ATG model by industry are presented in Table~\ref{tab:industry_wise_detail}.

\end{document}

%% file: figures/fig_proposed_dataset.tex
\begin{figure}[t]
 \centering
  \includegraphics[width=0.9\linewidth]{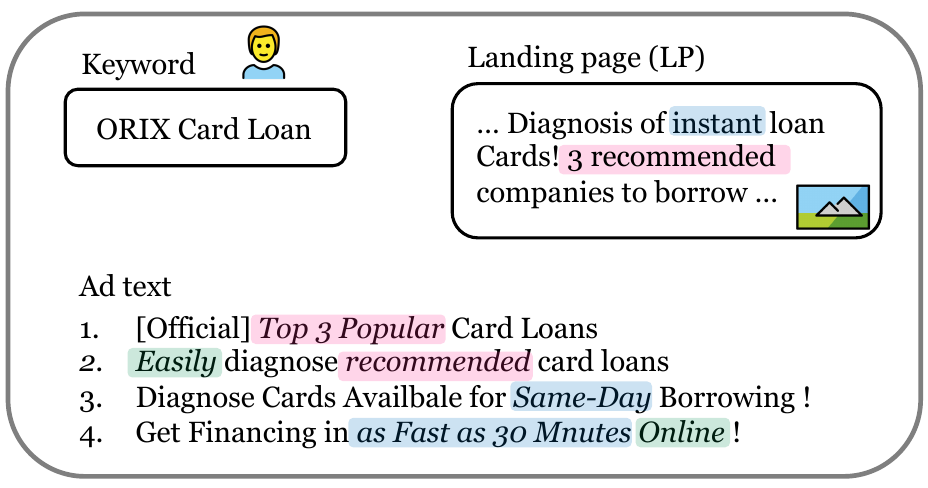}
\caption{Examples of our dataset, translated into English for visibility. The highlighted areas indicate the aspects of advertising appeals: \testbgfa{Speed}, \testbgfb{Trend}, and \testbgfc{User-friendliness}}\label{fig:camera_example}
\end{figure}

%% file: tables/tab_corpus_stats.tex
\begin{table}[t]
\centering
\small
\begin{tabular}{lrrr} \toprule
                      & \textbf{Train} & \textbf{Dev} & \textbf{Test} \\
                      \midrule
\# LP desc. \& user query & 12,395 & 3,098 & 872\\
\# reference per input          &   1    &  1    &  4 \\
\# tokens per an reference &            13.6      &     13.6         &    13.8 $\pm$ 0.7         \\
\# tokens per an LP desc. &          101.2      &     101.2         &     103.4          \\
\# tokens per an LP OCR  &      4649.6          &     4610.4         &    3510.3           \\
industry-wise        &                &              &     \checkmark \\        
\bottomrule
\end{tabular}\caption{Statistics of our dataset. Tokens are character units. \textit{\# tokens per an reference } in the test set shows the mean and standard deviation of the four references. \textit{Industry-wise} (\checkmark) indicates whether the data is separable by industry.}\label{tab:corpus_stats}
\end{table}

%% file: tables/tab_entity-type.tex
\begin{table*}[t]
\centering
\small
\begin{tabular}{lll} \toprule
\bf Entity type & \bf Input & \bf Output \\ \midrule
\textit{Time Expression}      &   2022年9月 （\textit{September 2022}）    &  2022年 (\textit{2022})   \\
\rowcolor{gray!7}
\textit{Katakana}    &  サイト (site)   &  ホームページ (homepage)     \\
\textit{Numerical Expressions}       &   6,800円 - 8,000円 (\textit{6,800 yen - 8,000 yen})  &  6,800円 (\textit{6,800 yen})  \\
\rowcolor{gray!7}
\textit{Named Entity}          &   イシダ (\textit{Ishida})    &  株式会社イシダ (\textit{Ishida Corporation})      \\
\textit{Terms}        &  求人情報 (\textit{Job Openings}) &  求人紹介 (\textit{Job Introductions})    \\ \bottomrule
\end{tabular}
\caption{The novel entity types used in our analysis and their corresponding examples. Katakana is a Japanese syllabary.}
\label{tab:entity_type}
\end{table*}

%% file: figures/fig_camera-train_hallucination.tex
\begin{figure}[t]
 \centering
  \includegraphics[width=0.9\linewidth]{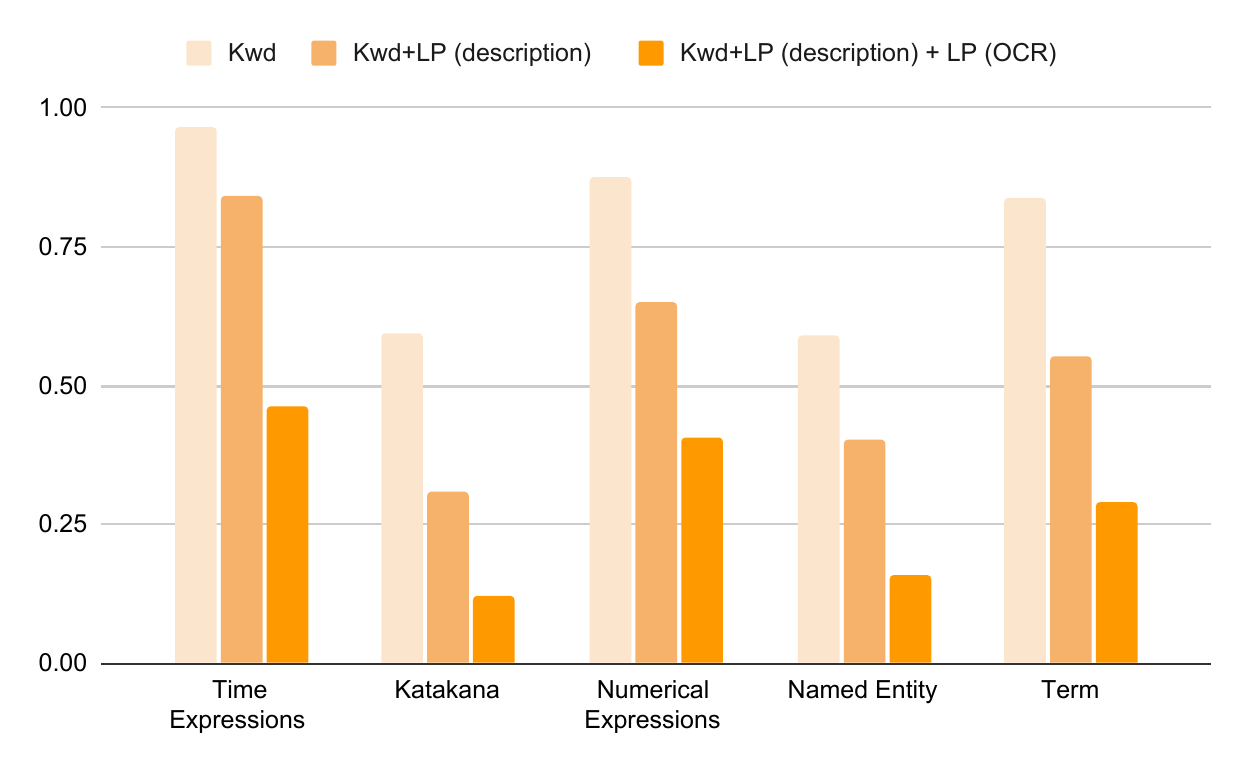}
 \caption{Percentages of novel entities included in our dataset when input information is increased.}
 \label{fig:camera-train_hallucination}
\end{figure}

%% file: tables/tab_agreement.tex
\begin{table}[t!]
\centering
\small
\tabcolsep=3pt
\begin{tabular}{lccc}
\toprule
          & \textbf{\small{Faithfulness}} & \textbf{\small{Fluency}} & \textbf{\small{Attractiveness}} \\ \midrule
All (= 3)      & 0.3    & 0.25   & 0.17    \\
Majority ($\geqq$ 2)     & -    & -    & 0.84    \\
\bottomrule
\end{tabular}
\vskip -2mm
 \caption{Inter annotator agreement.}
 \label{tab:agreement}
\end{table}

%% file: tables/tab_main_exp.tex
\begin{table}[t!]
\centering
 \small
\begin{tabular}{lrrr | rr}
\toprule
 & \multicolumn{1}{l}{B-4} & \multicolumn{1}{l}{R-1} & \multicolumn{1}{l}{BS} & \multicolumn{1}{l}{\textsc{Kwd}} & \multicolumn{1}{l}{\textsc{Reg}}  \\ 
\midrule
\multicolumn{5}{l}{\textbf{Unimodal model:}} \\
\midrule
BM25        & 5.4           & 16.1        &  70.1  & \textbf{97.0}    & 45.0      \\
BART        & \textbf{14.4} & 21.4        & 73.4       &   75.8    & 81.0     \\
T5          & 13.6          & \textbf{23.0} & \textbf{73.8}         & 89.8 & 78.5\\
GPT-3.5 & 3.5           & 14.2        &64.2          &    73.9  & 84.5     \\
GPT-4   & 4.4           & 16.4        & 65.1         &  78.6     & \textbf{87.0}    \\
Llama2 & 4.6           & 13.6        &  55.4         &  72.2   & 60.0   \\ 
\midrule
\multicolumn{4}{l}{\textbf{Multimodal models:}} \\
\midrule
T5 + \texttt{\{o\}} & \textbf{16.0}           & \textbf{24.7}        & \textbf{74.9}        &    \textbf{85.7}       & 70.0 \\
T5 + \texttt{\{o,l\}}   & 15.6   & 23.3        & 74.1        & 84.4   & 67.5 \\
T5 + \texttt{\{o,l,v\}} & 13.2    & 23.5        & 74.1      &  84.5 & \textbf{74.0}  \\ 
\bottomrule
\end{tabular}
\vskip -2mm
 \caption{Results: a \textbf{bold} value indicates the best result in each column.}
 \label{tab:main_results}
\end{table}

%% file: figures/fig_industry_wise_eval.tex
\begin{figure}[t]
 \centering
  \includegraphics[width=1.0\linewidth]{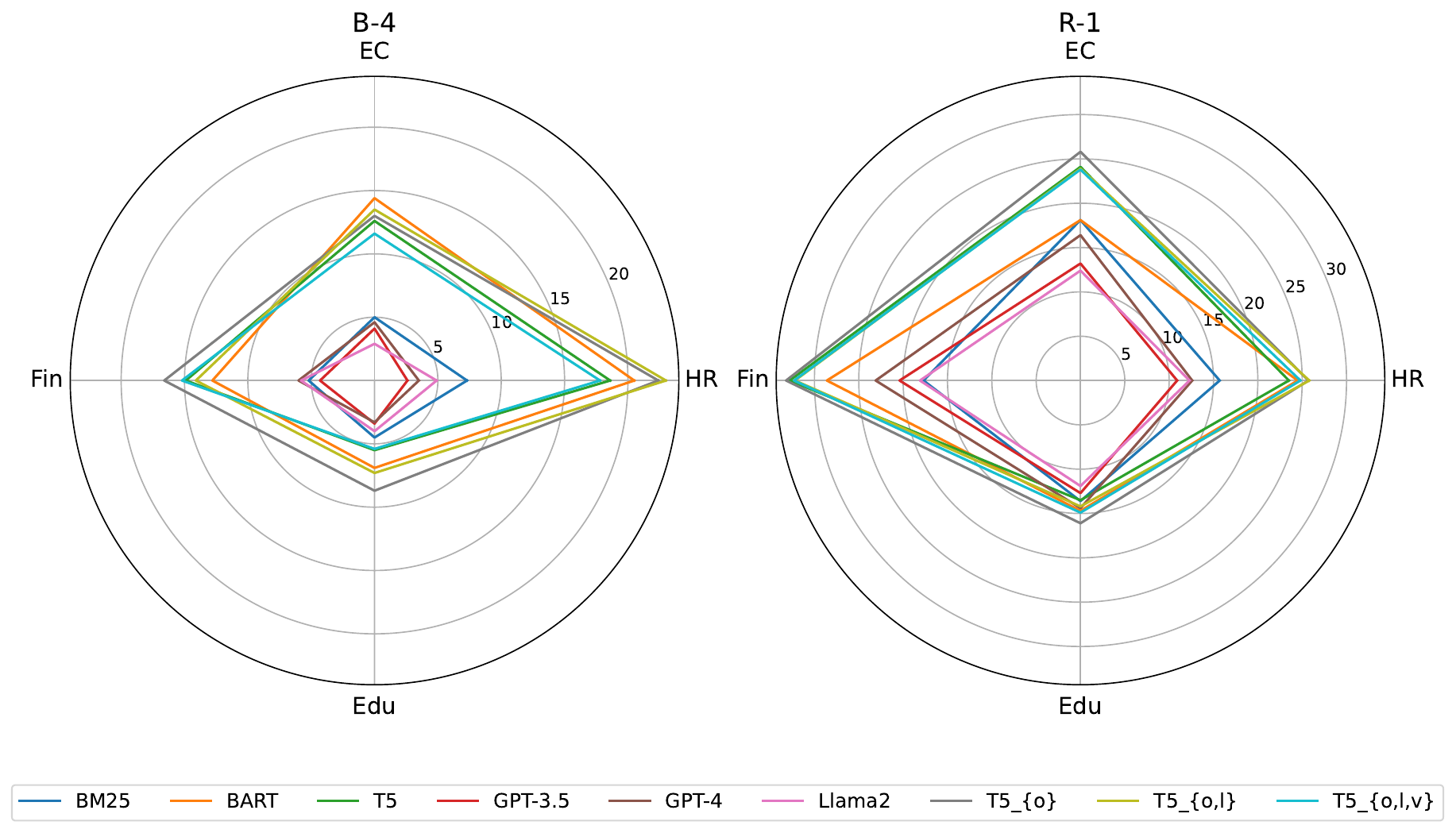}
 \caption{Industry-wise evaluation results.}
 \label{fig:industry_wise_eval}
\end{figure}

%% file: figures/fig_human-eval_faithful-fluency.tex
\begin{figure}[t]
    \begin{tabular}{c}
    
  \begin{minipage}[]{0.47\linewidth}
    \centering
    \includegraphics[width=1.0\linewidth]{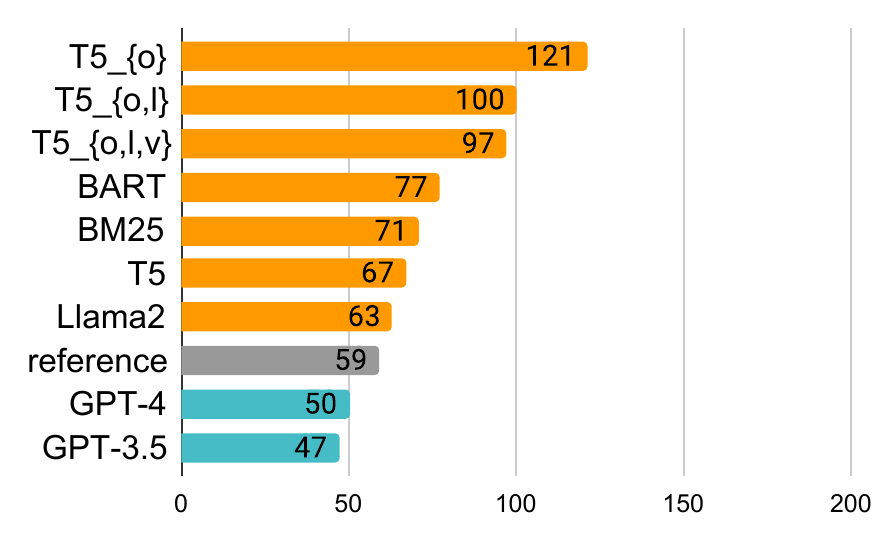}    
    \subcaption{Faithfulness}\label{fig:faithful}
  \end{minipage}

    \begin{minipage}[]{0.47\linewidth}
    \centering
    \includegraphics[width=0.9\linewidth]{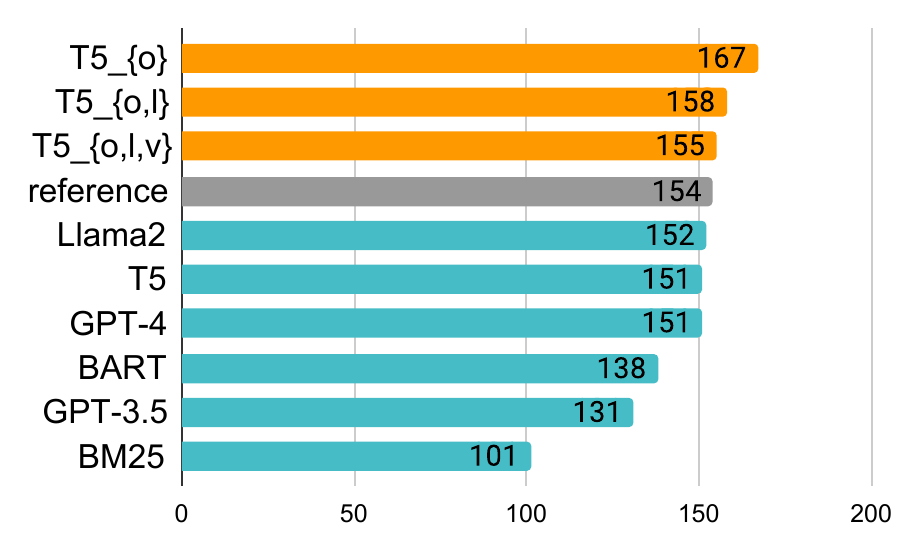}    
    \subcaption{Fluency}\label{fig:fluency}
  \end{minipage}

  \end{tabular}
  \caption{Human ranking in terms of faithfulness and fluency, respectively. }
  \label{fig:main_faithful_fluency}
\end{figure}

%% file: figures/fig_human-preference.tex
\begin{figure}[t]
 \centering
  \includegraphics[width=0.85\linewidth]{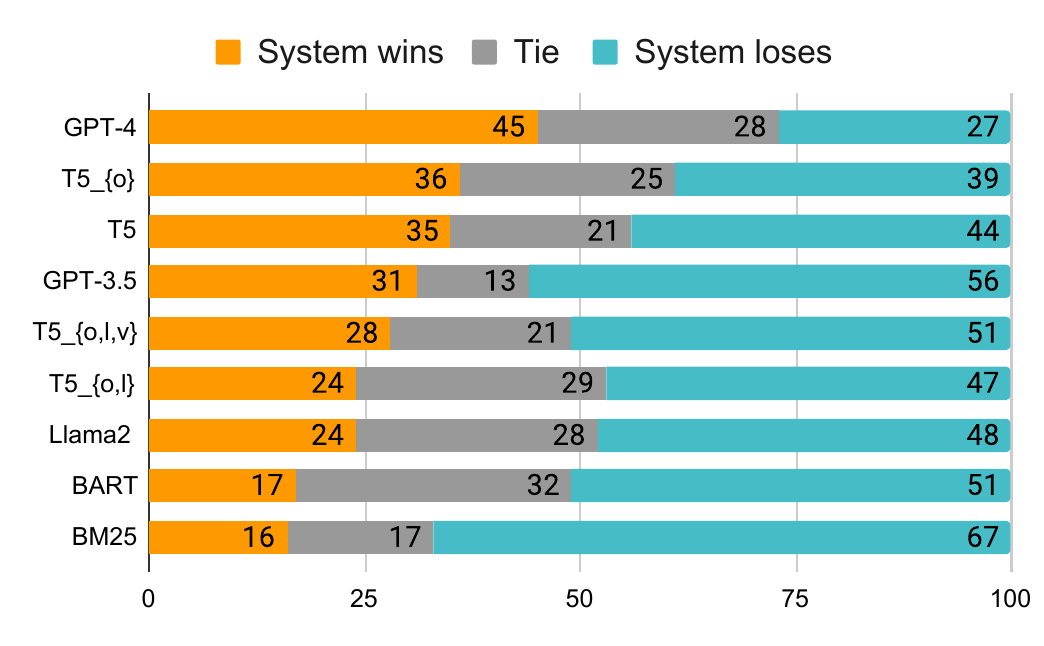}
 \caption{Human preference evaluation for each system output, comparing to a human-created reference.}
 \label{fig:attractivenss}
\end{figure}

%% file: figures/fig_multimodal_example.tex
\begin{figure}[t]
 \centering
  \includegraphics[width=0.9\linewidth]{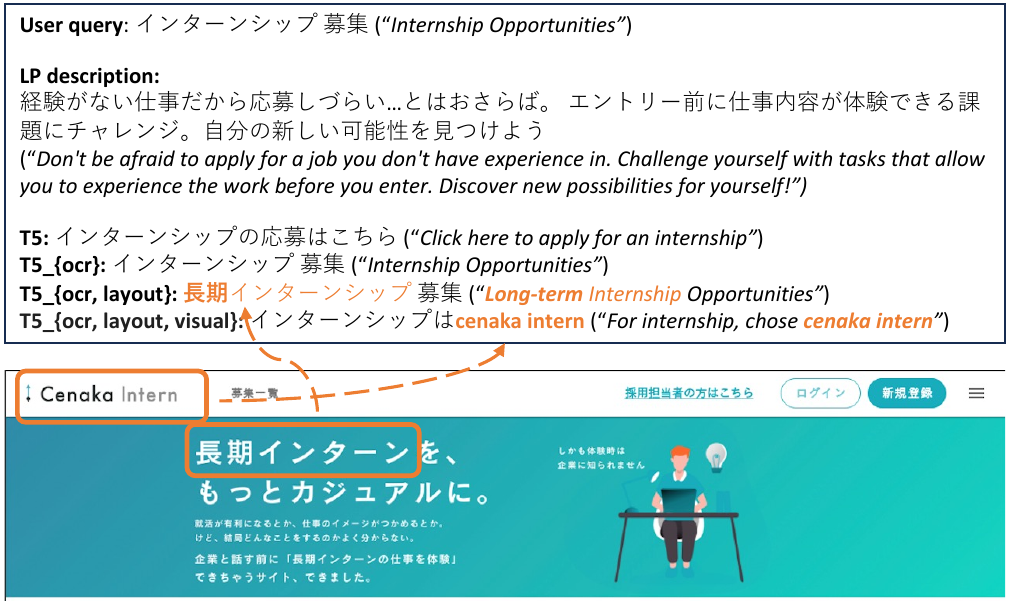}
 \caption{Example of how multimodal information in an LP contributed to the quality of the generated ad text.}
 \label{fig:multimodal_example}
\end{figure}

%% file: tables/tab_meta_eval.tex
\begin{table}[t!]
\centering
\small
\tabcolsep=3pt
\begin{tabular}{lcccccc}
\toprule
  & \multicolumn{2}{c}{\begin{tabular}{c} Faithfulness \end{tabular}} & \multicolumn{2}{c}{\begin{tabular}{c} Fluency \end{tabular}}  & \multicolumn{2}{c}{\begin{tabular}{c} Attractivenss \end{tabular}} \\
\cmidrule(r){2-3}\cmidrule(r){4-5}\cmidrule(r){6-7}
\multicolumn{1}{l}{Metrics}  & r  & $\rho$    & r  & $\rho$   & r  & $\rho$   \\
\midrule
B-4     & 0.88 & 0.83    & 0.53  & 0.30  & -0.12  & -0.68   \\
R-1       & 0.83 & 0.75    & \textbf{0.70}  & \textbf{0.55}   & \textbf{0.35}  & \textbf{0.03}     \\
BS      & \textbf{0.90} & \textbf{0.85}    & 0.67       &  0.50    & 0.20  & -0.20    \\
GPT-4       & 0.20 & -0.48    & -0.22       &  0.10   & -0.47  & -1.20      \\
\bottomrule
\end{tabular} 
\vskip -2mm
\caption{System-level meta-evaluation results with Pearson (r) and Spearman ($\rho$)}
 \label{tab:meta_eval}
\end{table}

%% file: figures/fig_kiwami_pref.tex
\begin{figure}[t]
 \centering
  \includegraphics[width=0.9\linewidth]{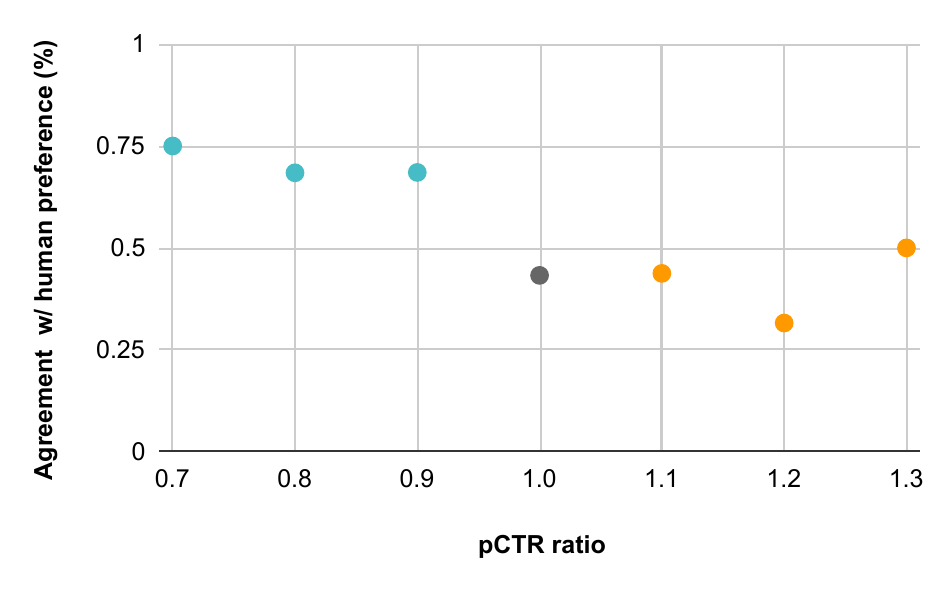}
 \caption{Agreement rate between human preference and pCTR.}
 \label{fig:kiwami_pref}
\end{figure}

%% file: figures/fig_example_sea.tex
\begin{figure}[t]
 \centering
  \includegraphics[width=1.0\linewidth]{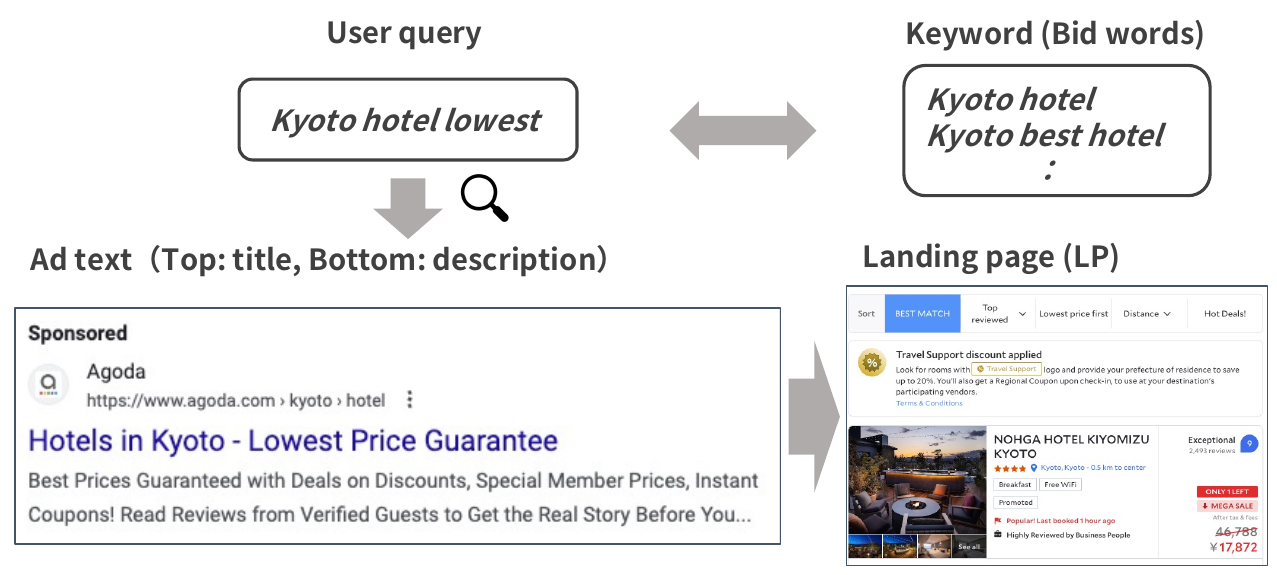}
 \caption{An example of search ads.}
 \label{fig:example_sea}
\end{figure}

%% file: tables/tab_exisiting_work_summary.tex
\begin{table*}[t]
\centering
\scriptsize
\begin{tabular}{lllllcc} \toprule
\bf Work           & \bf Approach & \bf Input        & \bf Output                    & \bf Affiliation & \bf Lang. & \bf xACL \\ \midrule
\citet{Bartz-et-al:08} &   Template       & Keyword       & Ad text  &    Yahoo          &  En     &      \\
\rowcolor{gray!7}
\citet{Fujita:10}    &   Template        & Promotional text     & Ad text, Keyword     &    Recruit       &  Ja     &      \\
\citet{Thomaidou-etal_snippet:13}      &    Template      & LP & Ad text      &   Athens Univ.           &   En    &      \\
\rowcolor{gray!7}
\citet{Hughes-et-al:19} &   Seq2Seq       & LP           & Ad text &     Microsoft         &   En    &      \\
\citet{Fukuda:19} &   Seq2Seq       & Keyword          & Ad text &     DENTSU         &   Ja    &      \\
\rowcolor{gray!7}
\citet{Mishra-et-al:20} &   Seq2Seq       & Ad text      & Ad text &      Yahoo        &  En     &      \\
\citet{Youngmann-etal:20}      &  Seq2Seq        & LP, Ad text      & Ad text &     Microsoft      &   En    &      \\
\rowcolor{gray!7}
\citet{Duan-etal:21}       &  Seq2Seq        & Query, KB  & Ad text        &     Tencent         &   Zh    &      \\
\citet{kamigaito-etal-2021-empirical}      &   Seq2Seq       &  LP, Query, Keyword    & Ad text &   CyberAgent           &  Ja     &    \checkmark   \\
\rowcolor{gray!7}
\citet{wang2021reinforcing}     &  Seq2Seq        &  LP,  Ad text    & Ad text &     Microsoft         &  En     &     \\ 
\citet{Zhang-etal:21}     &  Seq2Seq        & Ad text, Keyword, KB    &  Ad text &     Baidu         &  Zh     &     \\ 
\rowcolor{gray!7}
\citet{golobokov-etal-2022-deepgen}     &   Seq2Seq       & LP   & Ad text &   Microsoft           &   En    &  \checkmark   \\ 
\citet{kanungo-etal:22}     &    Seq2Seq      &  Multiple ad texts    & Ad text &    Amazon          &  En     &     \\ 
\rowcolor{gray!7}
\citet{wei-etal-2022-creater}     &    Seq2Seq      &  User review, Control code & Ad text &    Alibaba          &  Zh     &  \checkmark    \\ 
\citet{li-etal-2022-culg}     &    Seq2Seq      &  Query    & Ad text, Keyword &    Microsoft          &  En     &  \checkmark    \\ 
\rowcolor{gray!7}
\citet{Murakami-et-al:22}     &    Seq2Seq      &  Keyword, LP & Ad text &    CyberAgent          &  Ja     &      \\
\bottomrule
\end{tabular}
\caption{A summary of existing research on ad text generation. \textit{xACL} (\checkmark) presents whether the paper belongs to the ACL community, or some other research community (no \checkmark). }
\label{tab:summary_exisiting_work}
\end{table*}

%% file: tables/tab_example_ads_multimodal.tex
\begin{table*}[th]
\centering
\small
\begin{tabular}{lcl}
\toprule
LP desc. ($\bm{x}$)    & \multicolumn{1}{l}{user query ($\bm{a}$)} & ad text（$\bm{y}$）      \\ \midrule
\multirow{4}{*}{\begin{tabular}[c]{@{}l@{}}With our extensive service lineup and \\  dedicated professionals who are familiar \\  with your industry, we provide a  one-stop \\  solution to your recruiting needs.\end{tabular}} & 
\multirow{4}{*}{doda  enterprise}   & 1. To human resource managers - Doda Enterprise \\ 
&    & 2. For companies/ \testbgfa{Start using in as little as 1 day} \\
&  & 3. \testbgfb{One of the largest number of services in the industry}doda   \\
&    & 4. \testbgfb{Largest in the industry, boasting 7.08 million members.}  \\ 
\bottomrule                                                                                                                                                                    
\end{tabular}
\caption{An actual example of the difficulty of generating ad text without considering multimodal information of an LP, translated into English for visibility. The \testbgfb{red-highlighted} and \testbgfa{blue-highlighted} sections of the ad text have relevant information at the top (\ref{fig:lp_top}) and middle (\ref{fig:lp_middle}) of the LP, respectively.}\label{tab:example_ads_multimodal}

\end{table*}

%% file: figures/fig_example_LP.tex
\begin{figure*}[h!]
    \begin{tabular}{c}
    
  \begin{minipage}[]{0.45\linewidth}
    \centering
    \includegraphics[width=1.0\linewidth]{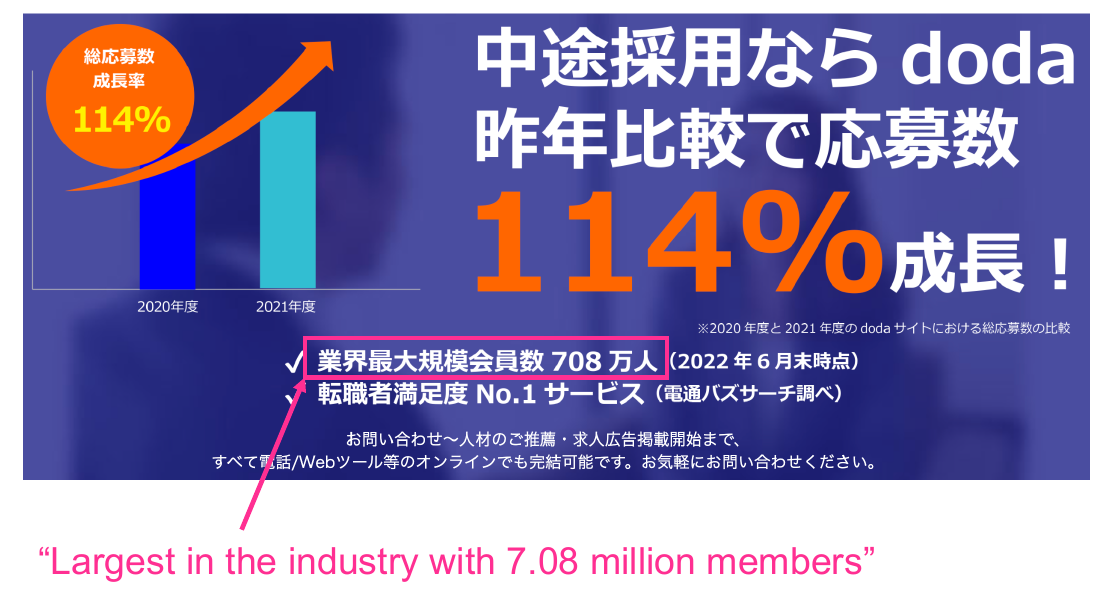}    
    \subcaption{LP (upper)}\label{fig:lp_top}
  \end{minipage}

    \begin{minipage}{0.06\hsize}
        \hspace{1mm}
    \end{minipage}
    \begin{minipage}[]{0.45\linewidth}
    \centering
    \includegraphics[width=1.0\linewidth]{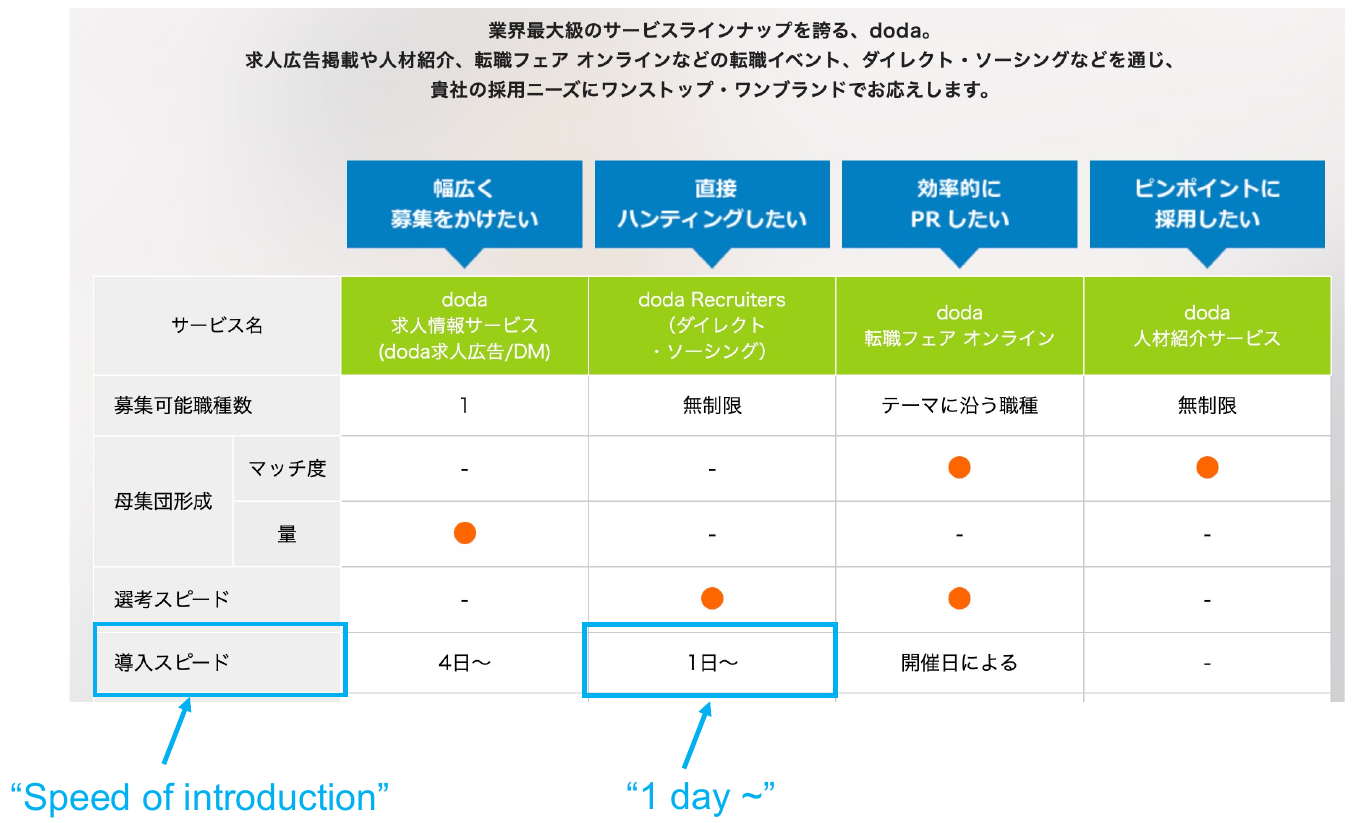}    
    \subcaption{LP (middle)}\label{fig:lp_middle}
  \end{minipage}

  \end{tabular}
  \label{fig:lp_example}
\end{figure*}

%% file: figures/fig_lp-to-text.tex
\begin{figure*}[t]
 \centering
  \includegraphics[width=0.85\linewidth]{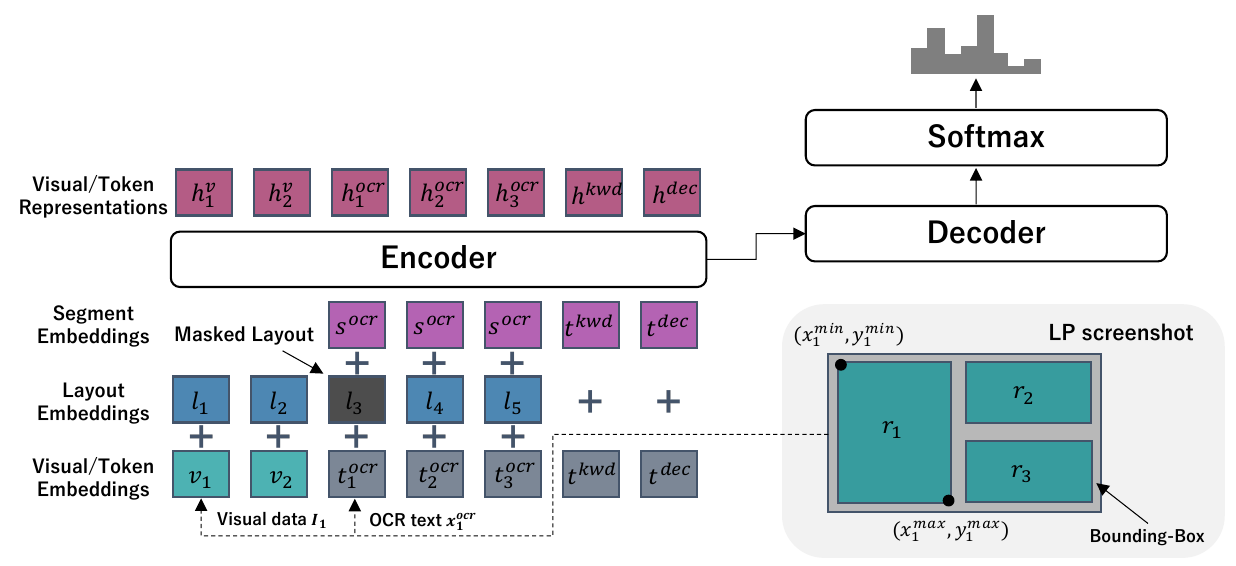}
 \caption{An overview of the model incorporating LP information, following \citet{Murakami-et-al:22}.}
 \label{fig:lp-to-text}
\end{figure*}

%% file: tables/tab_hyperparameters.tex
\begin{table*}[h!]
\centering
\begin{tabular}{@{}lp{100mm}@{}}
\toprule
 Hyperparameters & Values （BART / T5） \\ \midrule
 Models & \texttt{japanese\_bart\_base\_2.0} / \texttt{t5-base-japanese} \\
 Optimizer& Adam~\cite{kingma:2015:ICLR}\\
 Learning rate &  3e-4\\
 Max epochs & 20 \\
 Batch size & 8 \\
 Max length & 512 / 712 （T5+\texttt{\{o,l,v\}} only ） \\
 \bottomrule
\end{tabular}
\caption{Hyperparameters.}\label{tab:hyper_parameter}
\end{table*}

%% file: figures/fig_faithfulness.tex
\begin{figure}[t]
 \centering
  \includegraphics[width=1.0\linewidth]{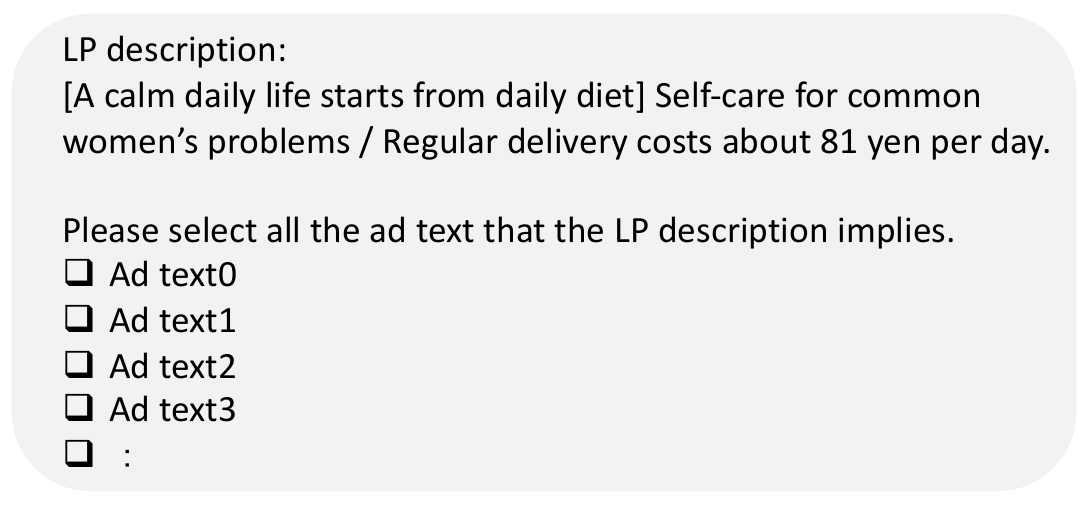}
 \caption{An example of annotation task in faithfulness evaluation.}
 \label{fig:faithul_anno}
\end{figure}

%% file: figures/fig_fluency.tex
\begin{figure}[t]
 \centering
  \includegraphics[width=1.0\linewidth]{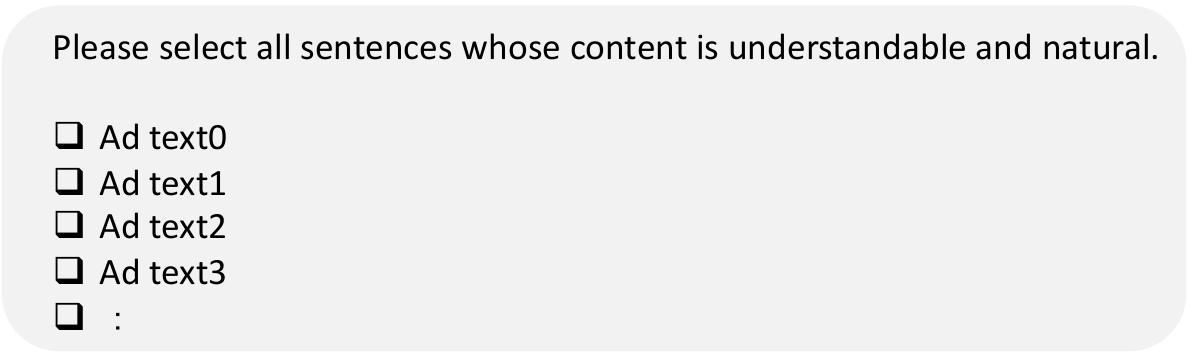}
 \caption{An example of annotation task in the fluency evaluation.}
 \label{fig:fluency_anno}
\end{figure}

%% file: figures/fig_attractivenss.tex
\begin{figure}[t]
 \centering
  \includegraphics[width=0.9\linewidth]{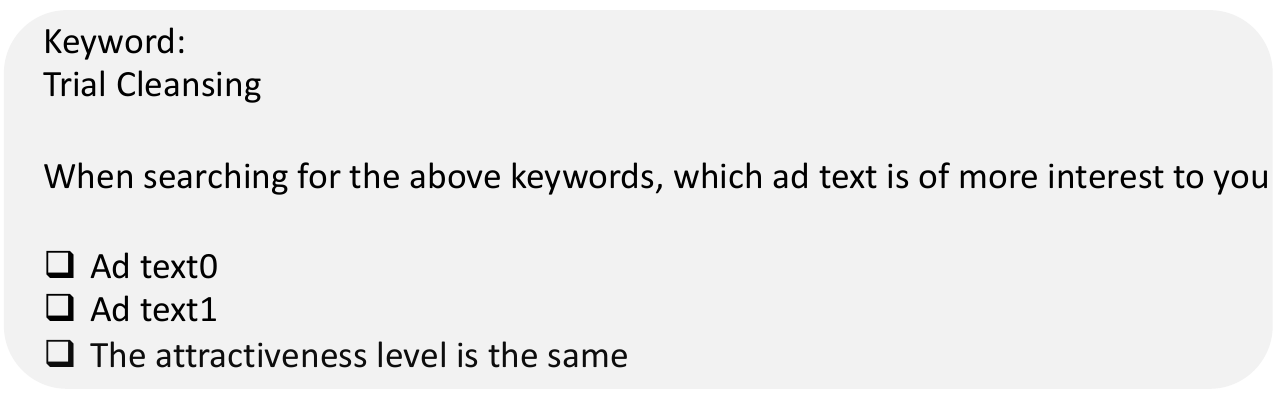}
 \caption{An example of annotation task in the attractiveness evaluation.}
 \label{fig:attractivenss_anno}
\end{figure}

%% file: tables/tab_industry-wise.tex
\begin{table*}[!h]
\centering
 \scriptsize
\begin{tabular}{l   rrrr rrrr rrrr rrrr}
\toprule
   & \multicolumn{4}{c}{\begin{tabular}{c} HR  \end{tabular}}& \multicolumn{4}{c}{\begin{tabular}{c} EC \end{tabular}} & \multicolumn{4}{c}{\begin{tabular}{c} Fin \end{tabular}} & \multicolumn{4}{c}{\begin{tabular}{c} Edu \end{tabular}} \\
\cmidrule(r){2-5}\cmidrule(r){6-9}\cmidrule(r){10-13}\cmidrule(r){14-17}
\multicolumn{1}{l}{Model}    & B-4 & R-1     & BS & \textsc{Kwd} & B-4 & R-1 & BS & \textsc{Kwd} & B-4 & R-1 & BS & \textsc{Kwd}& B-4 & R-1 & BS & \textsc{Kwd}\\
\midrule
\multicolumn{4}{l}{\textbf{Unimodal model:}} \\
\midrule
BM25   & 7.3  & 15.7 & 70.3 & \textbf{98.3} & 5.0 & 18.1 & 70.4& \textbf{98.3} & 5.2 & 17.7 & 70.3  & \textbf{99.0} & 4.5 & 13.6  & 69.5 & \textbf{93.3}\\
BART   & \textbf{20.5}  & \textbf{24.5} & 74.4 & 70.9 & \textbf{14.4} & 18.1 & 73.3  & 81.5 & 12.8 & 28.6 & 75.1 & 80.0  & \textbf{6.9} & \textbf{14.7}  & \textbf{81.0} & 73.0 \\
T5 & 18.6  & 23.5 & \textbf{74.7} & 84.8 & 12.6 & \textbf{24.1}  & \textbf{73.8} & 93.6 & \textbf{14.9} & \textbf{32.8} & \textbf{76.1} & 94.3 & 5.5 & 13.5 & 70.8 & 88.1\\ 
GPT-3.5  & 2.6  & 10.9 & 55.8 & 58.6 & 4.1 & 13.2   & 68.1 & 82.1 &4.3 & 20.4 & 69.4 & 85.7 & 3.4 & 12.7 & 65.5 & 72.6\\ 
GPT-4 & 3.5  & 12.6 & 56.0 & 65.4 & 4.6 & 16.4  & 68.2 & 85.5 & 6.0 & 23.1 & 71.5 & 89.0 & 3.3 & 14.4 & 66.2 & 77.4\\ 
Llama2  & 4.9  & 12.3 & 59.1 & 69.2  & 2.9 & 12.4 & 48.8 & 71.7 & 5.8 & 18.1 & 58.5  & 74.3 & 4.0 & 11.9 & 53.7 & 73.8 \\
\midrule
\multicolumn{4}{l}{\textbf{Multimodal model:}} \\
\midrule
T5 + \texttt{\{o\}}  & 22.4  & 25.7 & \textbf{75.5} & 82.3 & 13.0 & \textbf{25.8}   & \textbf{74.5} & \textbf{87.3} &\textbf{16.6} & \textbf{33.2} & \textbf{77.0} & 88.6 & \textbf{8.7} & \textbf{16.1} & \textbf{72.8} & \textbf{85.3} \\ 
T5 + \texttt{\{o,l\}} & \textbf{23.0}  & \textbf{25.8} & 74.9 & 81.4 & \textbf{13.5} & 23.9  & 73.7 & \textbf{87.3} & 14.1 & 32.2 & 76.2 & 86.2 & 7.3 & 14.2 & 71.8 & 83.7\\ 
T5 + \texttt{\{o,l,v\}}  & 17.8  & 24.8 & 74.4 & \textbf{82.7}  & 11.6 & 23.8 & 74.2  & 86.7 & 15.2 & 32.3 & 76.5  & \textbf{91.4} & 5.4 & 14.9 & 71.8 & 79.0 \\

\bottomrule
\end{tabular}
 \caption{Industry-wise evaluation results: a \textbf{bold} value indicates the best result in each column.}
 \label{tab:industry_wise_detail}
\end{table*}